%% file: main.tex
\documentclass[10pt,twocolumn,letterpaper]{article}

\usepackage{conf}
\usepackage{times}
\usepackage{epsfig}
\usepackage{graphicx}
\usepackage{amsmath}
\usepackage{amssymb}
\usepackage{ctable}
\usepackage{svg}
\usepackage{indentfirst}
\usepackage{caption}
\usepackage{enumitem}
\usepackage{multirow}

\DeclareMathOperator*{\argmax}{arg\,max}

\graphicspath{{figures/}}
\usepackage[pagebackref=true,breaklinks=true,letterpaper=true,colorlinks,bookmarks=false]{hyperref}

\conffinalcopy %

\def\confPaperID{} %

\begin{document}

\title{Joint Spatial and Layer Attention for Convolutional Networks}
\author{
	Tony Joseph*, \,\,
	Konstantinos G. Derpanis**, \,\,
	Faisal Qureshi*,\vspace{3pt}\\
    \{*University of Ontario Institute of Technology, **Ryerson University\}, Canada\\
	{\tt\small \{tony.joseph, faisal.qureshi\}@uoit.ca}
}

\maketitle

\begin{abstract}

In this paper, we propose a novel approach that learns to sequentially attend to different Convolutional Neural Networks (CNN) layers (i.e., ``what'' feature abstraction to attend to) and different spatial locations of the selected feature map  (i.e., ``where'') to perform the task at hand.  Specifically, at each Recurrent Neural Network (RNN) step, both a CNN layer and localized spatial region within it are selected for further processing. We demonstrate the effectiveness of this approach on two computer vision tasks: (i) image-based six degree of freedom camera pose regression and (ii) indoor scene classification.  Empirically, we show that combining the ``what'' and ``where'' aspects of attention improves network performance on both tasks. We evaluate our method on standard benchmarks for camera localization (Cambridge, 7-Scenes, and TUM-LSI) and for scene classification (MIT-67 Indoor Scenes). For camera localization our approach reduces the median error by 18.8\% for position and 8.2\% for orientation (averaged over all scenes), and for scene classification it improves the mean accuracy by 3.4\% over previous methods.
\end{abstract}
\section{Introduction} \label{sec:intro}

Convolutional Neural Networks (CNNs) \cite{lecun1990handwritten} are central models in a broad range of computer vision tasks, e.g., \cite{krizhevsky2012imagenet,He_2016_CVPR,he2017mask,ilg2017flownet,ledig2017photo}.  Generally, the processing of input imagery consists of a series of convolutional layers interwoven with non-linearities (and possibly downsampling) that yield a hierarchical image representation.
As deterministic processing proceeds in a CNN, both the spatial scope  (i.e., the effective receptive field) and the level of feature abstraction \cite{olah2017feature,zeiler2014visualizing} of the representation 
gradually increase.  Motivated by our understanding of human visual processing \cite{Rensink2000,tsotsos2011} and initial success in natural language processing \cite{bahdanau2014}, an emerging thread in computer vision research consists of augmenting CNNs with an attentional mechanism.  Generally speaking, the goal of attention is to dynamically focus computational resources on the most salient features of the input image as dictated by the task.

In this paper, we present an approach that incorporates attention into a standard CNN in two ways: (i) a layer attention mechanism (i.e., ``what'' layer to consider) selects a CNN layer, and (ii) a spatial attention mechanism selects a spatial region within the selected layer (i.e., ``where'') for subsequent processing.  Layer and spatial attention work in conjunction with a Recurrent Neural Network (RNN).  At each time step, first a layer is selected and next spatial attention is applied to it.

The RNN progressively aggregates the information from the attended spatial locations in the selected layers.  The aggregated information is subsequently used for regression or classification. Our model is trained end-to-end, without requiring additional supervisory labels. Empirically, we consider both regression (i.e., six degree of freedom, 6-DoF, camera localization) and classification (i.e., scene classification) tasks. 
Figure \ref{fig:overview} presents an overview of our approach to layer-spatial attention for 6-DoF camera localization. 

\input{intro.tex}

The guiding intuition behind our approach is that the optimal feature set for a task may be distributed across a variety of feature abstraction levels and spatial regions.  Here, we let an RNN identify the optimal features to aggregate.  For instance, in the context of image-based localization, a scene may contain both a set of salient objects captured by high-level features, such as a window or door, and  texture-like regions captured by low-level features.  Prior localization methods have exclusively relied on  either low-level features (e.g., \cite{Wang2006}) or high-level ones, e.g., \cite{anati2012,kendall2015}.  Our approach  considers the spectrum of feature abstractions in a unified manner.

\subsection{Contributions}

\noindent This paper makes the following contributions:
\begin{enumerate}[leftmargin=*]
\item We propose an attention model  
that learns to sequentially attend to different CNN layers (i.e., different
levels of abstraction) and different spatial locations (i.e., specific regions
within the selected feature map) to perform the task at hand.

\item We augment a standard CNN architecture, GoogLeNet \cite{szegedy2015}, with our attention model and empirically demonstrate its efficacy on
both regression and classification tasks: 6-DoF camera localization regression and indoor scene classification. We evaluate the proposed architecture on standard benchmarks: (a) Cambridge Landmarks, 7 Scenes, and TU Munich Large-Scale Indoor (TUM-LSI) for camera pose estimation; and (b) MIT-67 Indoor Scenes for scene classification. For camera localization our approach reduced the overall median error by 12.3\% for position and 13.9\% for orientation on Cambridge Landmarks, 19.3\% for position and 8.83\% for orientation on 7-Scenes, and 25.1\% for position and 1.79\% for orientation on TUM-LSI over the baseline \cite{walch2017}. For indoor scene classification on MIT-67 \cite{quattoni2009recognizing} our approach improves the mean accuracy by 3.4\% over the baseline \cite{hayat2016spatial}.  In both tasks, the baseline methods use the {\it same} base convolutional network.
\end{enumerate}

\begin{figure*}[t]
   \begin{center}
      \includegraphics[width=\textwidth]{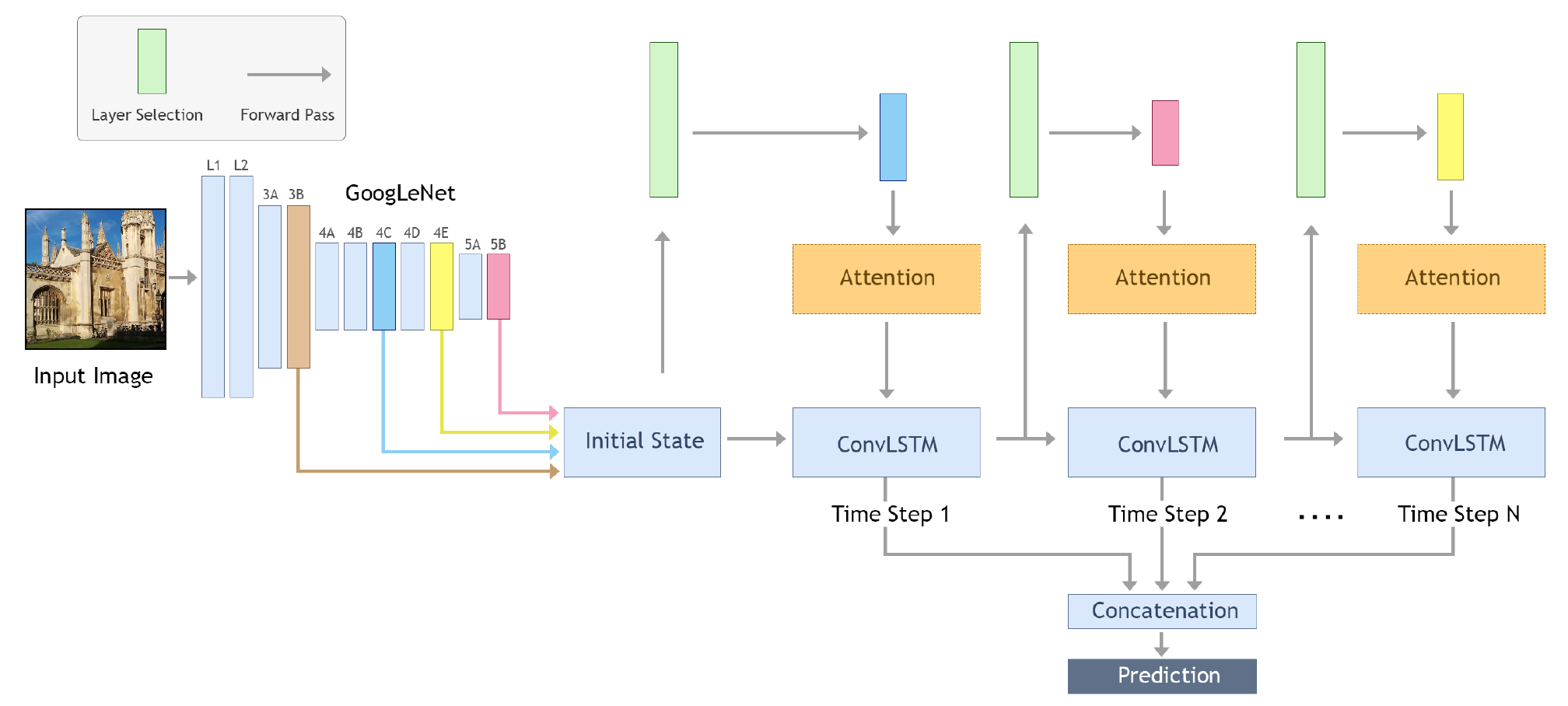}
   \end{center}
   \caption{Overview of our layer-spatial attention architecture. Layer-spatial attention is realized within a Conv-LSTM framework, where the layer attention uses the previous hidden state, and spatial attention uses both the selected layer and the previous hidden state. After $N$ Conv-LSTM steps, the hidden states from all steps are concatenated and used for regression or classification.}
   \label{fig:network}
\end{figure*}

\section{Related works}

\subsection{Attention.}
Attention is a mechanism that dynamically allocates computational resources to the most salient features of the input signal. Attention has  appeared in a variety of recent architectures
\cite{larochelle2010learning,zaremba2014recurrent,tang2014learning,ba2014,mnih2014,jayaraman2017,veit2018convolutional}. A natural way to implement a sequential attentional probing mechanism is with a Recurrent Neural Network (RNN) or variant (e.g., Long Short-Term Memory, LSTM \cite{hochreiter1997, zaremba2014recurrent}) in conjunction with a gating function \cite{StollengaMGS14, WangJQYLZWT17, xingjian2015convolutional} that yields a soft (e.g., softmax or sigmoid) or hard attention \cite{xu2015, williams1992}. The attentional policy is learned without an explicit training signal, rather the task-related loss alone provides the training signal for the attention-related weights. In this work, we incorporate both soft (spatial selection) and hard (layer selection) attention in an end-to-end trainable architecture.  Most closely
related to the current work are the soft and hard selection mechanisms proposed by Xu \etal \cite{xu2015} and Veit and Belongie \cite{veit2018convolutional}, respectively.
Xu \etal \cite{xu2015} proposed an end-to-end trainable soft spatial attention architecture for image captioning. We adapt this soft attention architecture for our purposes
and further extend it to include hard attention.  Veit and Belongie \cite{veit2018convolutional} proposed a dynamic convolutional architecture that selects whether or not information propagates through a given CNN layer during a forward pass.  Similar to Veit and Belongie \cite{veit2018convolutional}, we use the recently proposed Gumbel-Softmax to realize our discrete (hard) selection of layers. 

\subsection{Image-based camera pose localization.}
Low-level features (e.g., SIFT \cite{lowe2004}) have dominated the camera pose localization literature, e.g., \cite{anati2012, sattler2017, li2012worldwide, engel2014lsd}.  An early example of using high-level features for camera localization appeared in Anati \etal\ \cite{anati2012}, where heatmaps from object detections were used for localization. More recently, high-level CNN features have garnered attention. These features can be considered as soft proxies to object detections. Kendall \etal \cite{kendall2015, kendall2016} proposed PoseNet, an image-based 6-DoF camera localization method.  PoseNet regresses the camera position and orientation based on input provided by a CNN layer Kendall and Cipolla \cite{kendall2017geometric} reconsidered the loss used in PoseNet to integrate additional geometric information. Walch \etal  \cite{walch2017} extended the PoseNet approach by introducing an LSTM-based dimensionality reduction step prior to regression to avoid overfitting. In each case, the networks rely on features from a manually selected layer, located relatively high in the feature hierarchy.  In contrast, we propose an attentional network that is capable of dynamically integrating the most salient features across the spectrum of feature abstractions (capturing potentially texture-like and object-related features as necessary). 

\subsection{Indoor scene classification.}
To demonstrate the generality of our approach we also consider a classification task, indoor scene classification.  Here, a wealth of research has considered both handcrafted (e.g., \cite{doersch2013mid,juneja2013blocks}) and learned deep features, e.g., \cite{sharif2014cnn,hayat2016spatial}. In this work, we compare our approach using a standard deep architecture, GoogLeNet \cite{szegedy2015}, which we also use as the base network for our layer-spatial attention method.

\section{Technical approach} \label{uan}

Our layer-spatial attention network sequentially probes the input signal over a fixed number of steps. It is comprised of a hard selection mechanism that selects a CNN layer (Sec.\ \ref{what}) and soft attention that selects a spatial location within the selected layer  (Sec.\ \ref{where}). The attention network is realized using a convolutional LSTM (Conv-LSTM) \cite{xingjian2015convolutional}. Figure \ref{fig:network} provides an overview of our architecture. At each Conv-LSTM step, the layer attention selects a CNN layer and spatial attention localizes a region within it.  After $N$ recurrent steps, the Conv-LSTM hidden states for all steps are concatenated and used for classification or regression.

\subsection{Where: Spatial attention} \label{where}
We adapt the recurrent model from Xu \etal \cite{xu2015} with soft spatial attention as the foundation of our method.  At each time step $t$, the spatial attention
mechanism receives as input the selected layer $\mathbf{f} \in \mathbb{R}^{h_f \times w_f \times d_f}$ (see Sec.\ \ref{what}) and the recurrent hidden state $\mathbf{h}_{t} \in \mathbb{R}^{h_h \times w_h \times d_h}$ from the previous step.
The soft attention layer is implemented as follows:
\begin{align}
\mathbf{h}_{att}  &= \mathbf{h}_{t} * \mathbf{E}_h \nonumber \\
\mathbf{f}_{att}  &= \mathrm{ReLU}(\mathbf{h}_{att} + \mathbf{f}) \label{softattn_eq} \\  
\mathbf{O}_{att}  &= \mathrm{softmax}(\mathbf{f}_{att} * \mathbf{C}_A) \odot \mathbf{f} , \nonumber
\end{align}
where $*$ denotes the convolutional operator and $\odot$ is element-wise multiplication. The attention layer consists of two convolutional layers,  $\mathbf{E}_h$ and $\mathbf{C}_A$, which compute an embedding and (unscaled) attention mask, respectively. The embedding layer,  $\mathbf{E}_h$, is used to transform the hidden state channel  dimension to bring it equal to the input layer's channel dimension. The $\mathbf{C}_A$ layer computes the unscaled attention mask with dimensions $h_f \times w_f \times 1$. The final attention mask is computed by taking the $\mathrm{softmax}$ of the unscaled attention mask. The output of the attention layer $\mathbf{O}_{att}$ is obtained by taking an element-wise multiplication between the features in each channel and attention map.

\subsection{What: Layer attention} \label{what}
In layer attention (i.e., ``what'' features to attend) a CNN layer is selected whose feature map is deemed to contain the most salient information at the current recurrent step.  Our layer attention involves a discrete (hard) selection of a CNN layer. Here, we use the recently proposed continuous relaxation of the Gumbel-Max trick \cite{gumbel}, the Gumbel-Softmax \cite{maddison2014sampling,jang2016categorical}, to realize the discrete selection of layers.

Gumbel-Max provides a simple and efficient way to draw samples from a categorical (discrete) distribution:
\begin{equation}
   z = \mathrm{one\_hot}(\argmax[g_i + \log \pi_i ]), \label{eq3}
\end{equation}
where, $g_1,..., g_k$ are i.i.d.\ samples drawn from the $\mathrm{Gumbel(0, 1)}$ distribution, and $\pi_i$ are unnormalized probabilities.  Samples $g$ are drawn using the following procedure: (i) draw sample $u \sim \mathrm{Uniform(0, 1)}$; and (ii) set $g = -\log(-\log(u))$. In the forward pass (and during testing), we compute the $\text{arg max}$ of the unnormalized probabilities.  In contrast, in the backward pass the $\argmax$ is approximated with a softmax function:
\begin{equation}
   y_i =  \frac{\exp  \left ( \frac{\log \left (\pi_i \right) + g_i}{\tau} \right)}
               {\sum_{j=1}^{k} \exp  \left ( \frac{\log \left (\pi_j \right) + g_j}{\tau} \right)}, \label{eq4}
\end{equation}
where $k$ is the number of CNN layers that are considered for selection, $i \in [1,k]$, and $\tau$ represents temperature. (This approach is the straight-through version of the Gumbel-Softmax estimator proposed in \cite{jang2016categorical}.) During training the temperature, $\tau$, is progressively lowered.  As the temperature approaches zero, samples from the Gumbel-Softmax distribution closely approximate those drawn from a categorical distribution.

For layer attention, we realize the (layer) selection scores (i.e., unnormalized probabilities) at each recurrent step as the output of a fully connected layer computed using the previous hidden state. During the forward pass we perform layer selection using Eq.\ \ref{eq3} and in the backward pass gradients are computed using Eq.\ \ref{eq4} 
to keep our architecture end-to-end trainable.

\subsection{Tasks} \label{sec:training}

In our approach, after $N$ Conv-LSTM steps, the hidden states are concatenated, average pooled, and passed onto a fully connected layer for (regression/classification) prediction. To ensure that our comparisons are meaningful, and that any differences in the performance of our method to those posted by previous methods are due to our attention mechanism, we use the exact same losses as those used by our baselines. 

\begin{figure*}[t]
   \begin{center}
      \includegraphics[width=\textwidth]{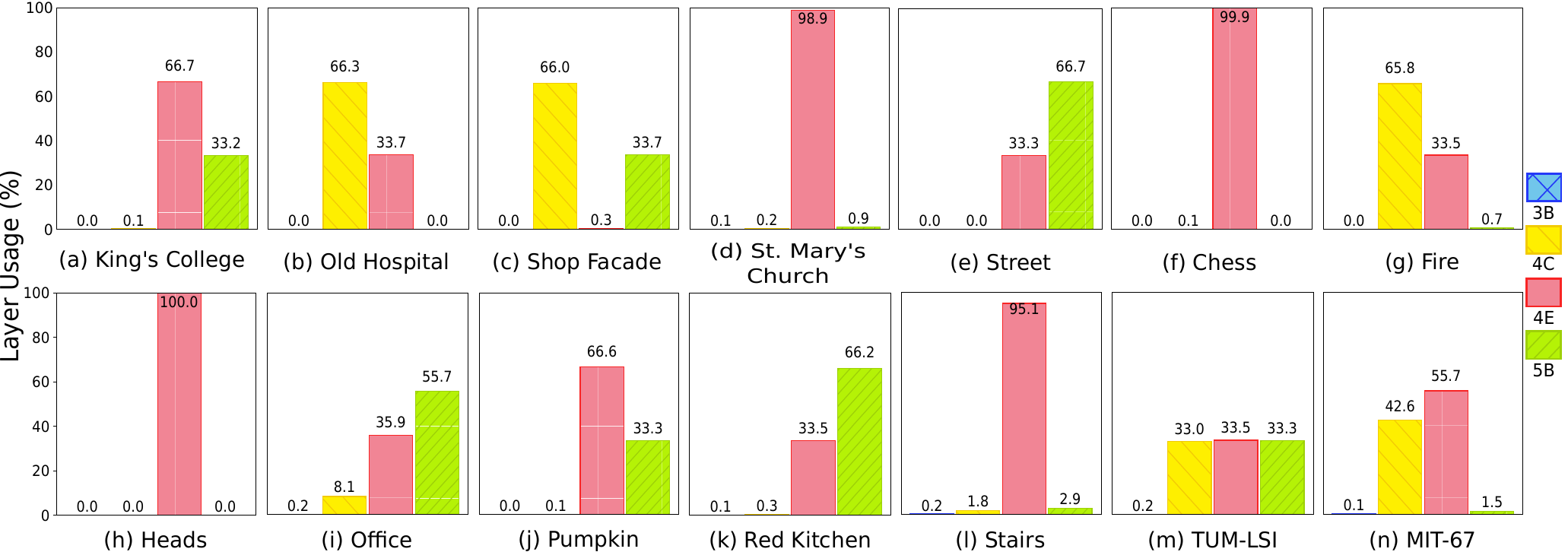}
   \end{center}
   \caption{Layer Selection Frequencies (LSF) on all four datasets on the test set. (a) - (e) are Cambridge Landmarks scenes, (f) - (l) are scenes from 7-Scenes, (m) and (n) are TUM-LSI, and MIT-67 dataset, respectively. The bins refer to the GoogLeNet\cite{szegedy2015} Conv-\{3B, 4C, 4E, 5B\} layers.  The vertical axis represents layer usage percentages.}
   \label{fig:lsf}
\end{figure*}

\subsubsection{Camera pose estimation} 
The proposed camera localization network takes an RGB image as input and outputs camera position and orientation $[\hat{\mathbf{x}}, \hspace{5pt} \hat{\mathbf{q}} ]^\top$. Camera pose is defined relative to an arbitrary reference frame. We use the same regression loss as our baselines \cite{kendall2015, kendall2017uncertainties, walch2017} to facilitate
direct empirical comparison:
\begin{equation}
   \mathcal{L} = \| \mathbf{x} - \hat{\mathbf{x}} \|_2 +
    \beta \|\mathbf{q} - \frac{\hat{\mathbf{q}}}{\|\hat{\mathbf{q}}\|_2}\|_2, \label{eq6}
\end{equation}
where $[\mathbf{x} ,\hspace{5pt} \mathbf{q}]^\top$ represent ground truth position $\mathbf{x}$ and orientation $\mathbf{q}$, and $[\hat{\mathbf{x}}, \hspace{5pt} \hat{\mathbf{q}} ]^\top$ denote predicted position $\hat{\mathbf{x}}$ and orientation $\hat{\mathbf{q}}$.  Orientations are represented using quaternions. $\beta$ is a scalar hyperparameter that determines the relative weighting between the positional and orientation errors. We use the same $\beta$ value as our baselines, PoseNet \cite{kendall2015} and LSTM-PoseNet  \cite{walch2017}. 

\subsubsection{Indoor scene classification}
Consistent with our scene classification baseline \cite{szegedy2015},
we use the standard cross-entropy classification loss:
\begin{equation}
  \mathcal{L} = - \mathbf{y}_{c}^\top \log(\hat{\mathbf{y}}_{c}), \label{eq5}
\end{equation}
where $\mathbf{y_{\mathrm{c}}}$ is a one-hot encoded class label for class $c$, and $\hat{\mathbf{y}}_{\mathrm{c}}$ is the output of the softmax classifier.

\subsection{Implementation details}

To realize our layer-spatial attention model we use the same basic architecture as Xu \etal \cite{xu2015} for sequential spatial attention.  We augment this 
 network with hard attention for layer selection.  To avoid overfitting,
 we replace the LSTM layers with ConvLSTM \cite{xingjian2015convolutional} layers 
that reduce the network weight parameterization.  The hidden state size is set to $96$. In this work we used a multi-convolutional layer modeled after the Inception module \cite{szegedy2015} for layer-spatial selection; see the supplementary material for further discussion. All experiments use GoogLeNet \cite{szegedy2015} as the feature extractor to maintain meaningful comparisons with the baseline methods. It is conceivable that using a different base network may yield improved results; however, the focus of our experiments is to study the impact of our proposed layer-spatial attention mechanism.

For practical reasons we selected a sparse set of layers (Conv-\{3B, 4C, 4E, 5B\}) that capture a range of abstractions. It is straightforward to extend the network to select any layer; however, it will considerably increase the training time.  Another consideration is that the layers often have different channel dimensions, which necessitates additional weights for embedding layers. All models were trained end-to-end using the ADAM \cite{kingma2014adam} optimizer. Batch-Norm \cite{ioffe2015batch} with default parameters is applied to both spatial attention and layer selection network.  Our code is implemented using TensorFlow 1.4 \cite{abadi2016tensorflow}. Additional details about our architecture are provide in the supplementary material.

\section{Empirical evaluation}

\subsection{Datasets} 

We evaluate our layer-spatial attention model on a
variety of standard datasets.  For 6-DoF camera localization we evaluate on  
Cambridge Landmarks~\cite{kendall2015}, 7-Scenes~\cite{shotton2013scene}, and TUM-LSI~\cite{walch2017}.  For scene classification we evaluate on MIT-67 Indoor Scenes \cite{quattoni2009recognizing}. (Additional information on these datasets can be found in the supplementary materials.) For camera pose estimation, we resize the images to $256\times455$ pixels. 
As done in our localization baselines \cite{kendall2015,walch2017}, separate mean images are computed for each colour channel and the images are mean subtracted per channel. 
For indoor scene classification, we resize the images to $256\times256$. 
For all experiments, we use crops of $224\times224$ pixels (random crops during training and center crops during testing). For indoor scene classification we also used random horizontal flips during training.

\subsection{Results}
\input{camera-pose-table.tex}

\input{scene-classification-table.tex}

\begin{figure*}
    \begin{center}
    \includegraphics[width=\linewidth]{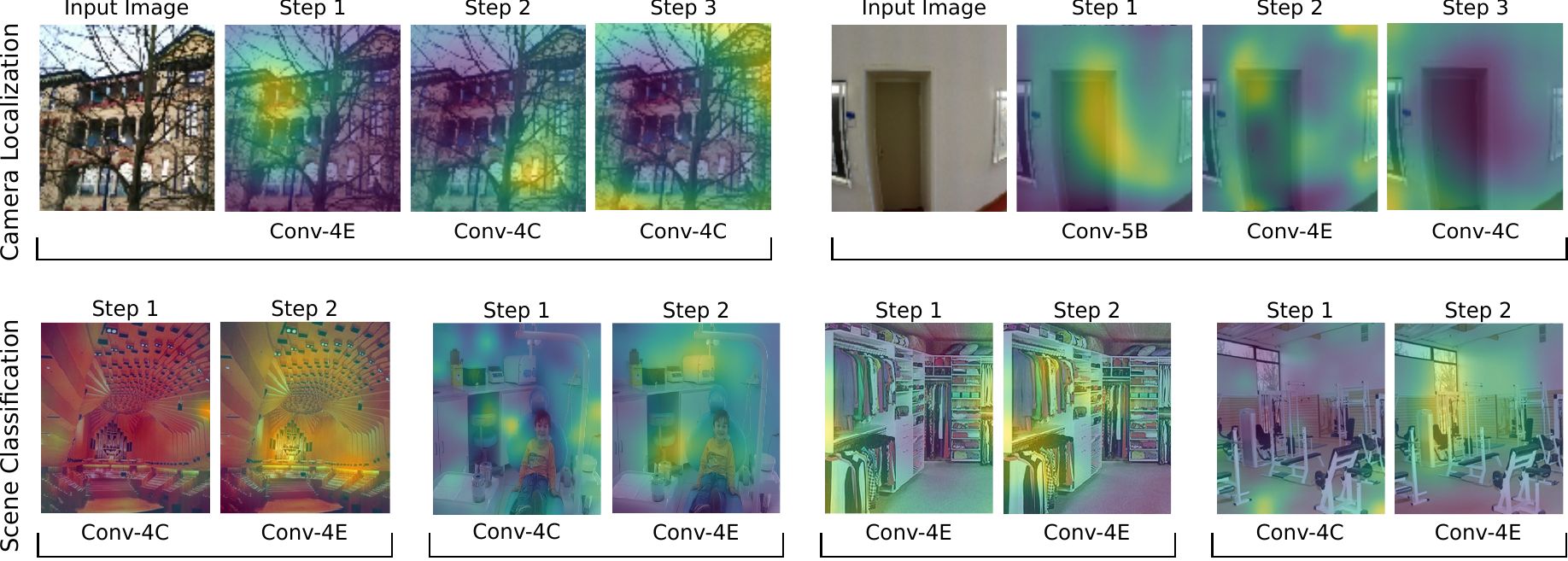}
    \end{center}
    \caption{Qualitative results on camera pose localization (top row) and indoor scene classification (bottom row). Top row: input image along with the spatial attention superimposed on the input image for three Conv-LSTM steps. Bottom row: spatial attention superimposed on the input image for two Conv-LSTM steps. The labels underneath each image indicates the selected CNN layer.}
    \label{fig:qualitative}
\end{figure*}

\noindent Figure~\ref{fig:lsf} shows the frequencies of the GoogLeNet feature layers selected for each dataset on the respective test sets.  As can be seen, the datasets predominately utilize more than one layer.  Furthermore, the layers most frequently selected differ widely amongst the datasets. We found that for image-based camera localization using three Conv-LSTM steps worked best, after which the performance decreases, the error increases. In the case of indoor scene classification two Conv-LSTM steps performed best. Additional experiments using five recurrent steps are shown in the supplementary materials for  both tasks.

\noindent \textbf{Camera localization.} \space Table ~\ref{tbl:pose-results-cam} compares our proposed method against  image-based camera pose regression methods \cite{kendall2015,kendall2016,walch2017}. All the compared methods use GoogLeNet as the source of features for regression, with the baselines limiting features to layer Conv-5B. In terms of the individual scenes, our method achieves the least error in both translation and rotation in the majority of cases at three steps. Considering the aggregate results over the respective datasets, we see our method yields significant improvements over the state-of-the-art, ranging between $12.3$ and $25.1$ percent for translation and $1.79$ and $13.9$ percent for rotation.

The TUM-LSI  dataset contains large textureless surfaces and repetitive scene elements covering over $5,575$ $m^2$.  Active search or SIFT-based approaches have been previously shown to perform poorly on this dataset \cite{walch2017}.  Our method achieves state-of-the-art performance, suggesting that the ability to attend to different CNN layers over successive LSTM steps helps. Figure \ref{fig:qualitative} (top row) shows qualitative results for camera localization. For outdoor scenes, it appears our attention mechanism captures both low-level (e.g., corners) and high-level structures (e.g., rooftops and windows).

\noindent \textbf{Indoor scene classification.} \space Table \ref{tbl:scene-results} compares our proposed layer-spatial attention method against three baselines \cite{sharif2014cnn,hayat2016spatial,szegedy2015}. The proposed method achieves best performance after two recurrent steps. Figure \ref{fig:qualitative} (bottom row) shows several qualitative results for indoor scene classification. The layer-spatial attention seems to capture objects and physical scene structures present in the scene. For the Concert Hall image, the attention mechanism appears to focus on the entire image, perhaps focusing on the scene architecture. For the Dental Office image, spatial attention picks out the dental equipment (a permanent fixture) and correctly ignores the person (a transient entity). For the Closet image, clothes and cabinetry are selected. Finally, for the Gym image, the proposed attention mechanism selects the exercise equipment. 

\subsection{Ablation study}

\input{ablation-study-table.tex}

Table~\ref{tab:ablation} summarizes an ablation study that we performed to gauge the impact of combining layer selection with spatial attention. We choose Old Hospital (Cambridge Landmarks), Office (7-Scenes), TUM-LSI, and MIT-67 datasets for this ablation study.  Old Hospital and Office were selected since we found these to be the most challenging for our proposed network.

We manually selected GoogLeNet's Conv-\{3B, 4E, 5B\} layers and applied spatial attention to each independently.  (Note, the PoseNet results reported in Table \ref{tbl:pose-results-cam} use layer Conv-5B without any form of attention for direct position-orientation regression.)
Our results confirm that it is sometimes beneficial to use layers other then the final CNN layer.  Median localization errors, for example, improve for both Old Hospital and Office datasets when we use layers other than Conv-5B. 
Note that in previous camera pose localization works \cite{kendall2015,kendall2016,walch2017} Conv-5B was manually selected. 
For indoor scene classification, selecting Conv-4E yields the best result.
The last column of Table~\ref{tab:ablation} includes results obtained by combining layer selection and spatial attention.  Notice that in three out of four cases shown, network achieves best performance (lowest errors in case of camera pose estimation, and highest accuracy in case of indoor scene classification) when using both layer selection and spatial attention.  The second last column in Table~\ref{tab:ablation} includes results when using layer selection alone.  The network performance deteriorates when spatial attention is absent.  

Our results are consistent with our initial guiding intuition that salient information is distributed across the spectrum of feature abstractions, e.g., things vs.\ stuff. Our proposed layer-spatial attention mechanism exploits this aspect to achieve better performance.

\section{Conclusion}
In this paper, we have presented an architecture that dynamically probes  the convolutional layers of a CNN  to aggregate and process the optimal set of features for a given task.  We introduced an attention architecture that learns to sequentially attend to different CNN layers (i.e., levels of feature abstraction) and different spatial locations within the selected layer. In the context of two vision tasks, camera localization and scene classification, we  empirically showed that our approach to layer-spatial attention improves regression and classification performance over manually selecting layers and previous approaches. Our proposed approach to attention is general and may prove useful for other vision tasks.

\section{Acknowledgments}
The authors would like to thank Kamyar Nazeri (Imaging lab, University of Ontario Institute of Technology, Canada) for providing support on figures presented in this paper. We gratefully acknowledge the support of NVIDIA Corporation with the donation of the Titan Xp GPU used for this research.

{
\bibliographystyle{ieee}
\bibliography{mybib}
}

\clearpage
\newpage
\appendix
\onecolumn
\title{Supplementary Materials}

\maketitle

\section{Detailed Attention Architecture} \label{appendix:uan_arch}

\begin{figure}[!h]
\centering
\includegraphics[width=\textwidth]{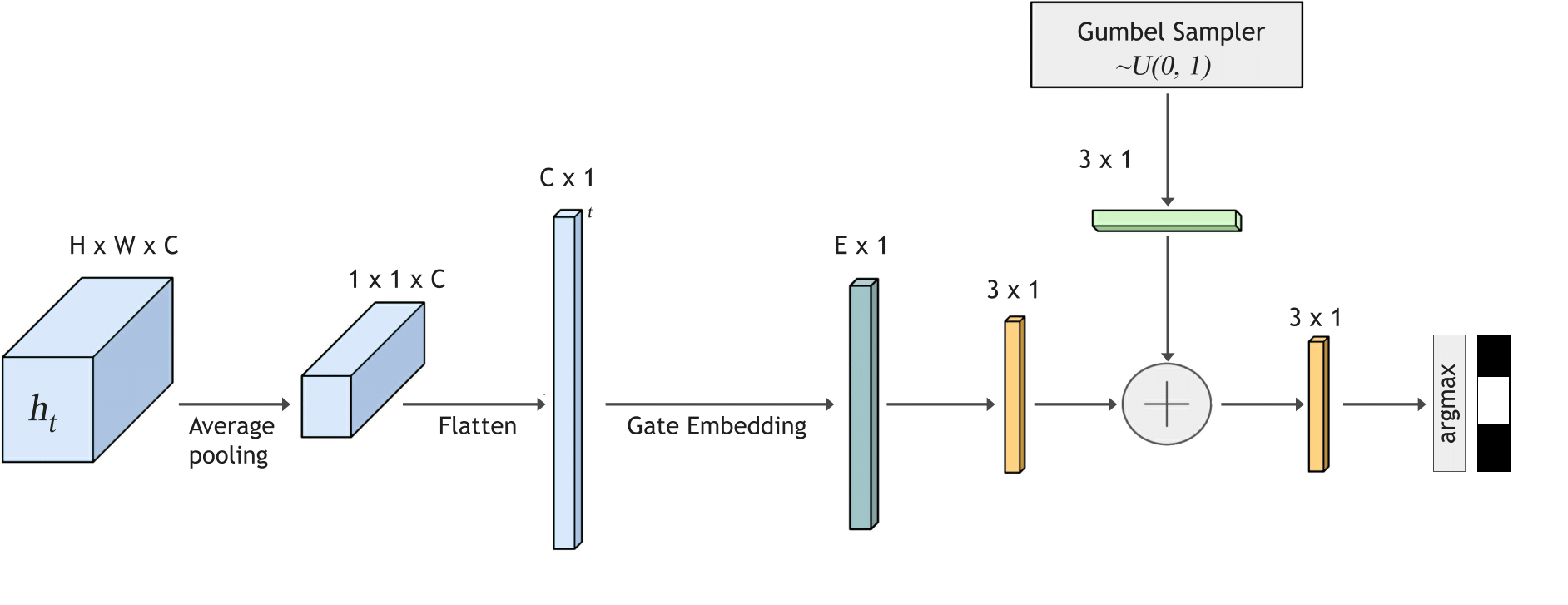}
\caption{Layer Selection Mechanism.}
\label{fig:select_select}
\end{figure}

Figure~\ref{fig:select_select}, illustrates the layer selection mechanism. The mechanism receives input $\mathrm{h}_t$ from ConvLSTM. It then performs an average pool and an intermediate gate embedding before prediction. We add the Gumbel samples to the predicted logits and perform an $argmax$ to select the optimal layer. The gate embedding layer dimension $\mathrm{E}$ is much smaller than $\mathrm{C}$. This gate embedding layer helps build a possible representation of incoming features at every LSTM steps, without significantly increasing the network parameters.

\begin{figure}[t]
\includegraphics[width=\textwidth]{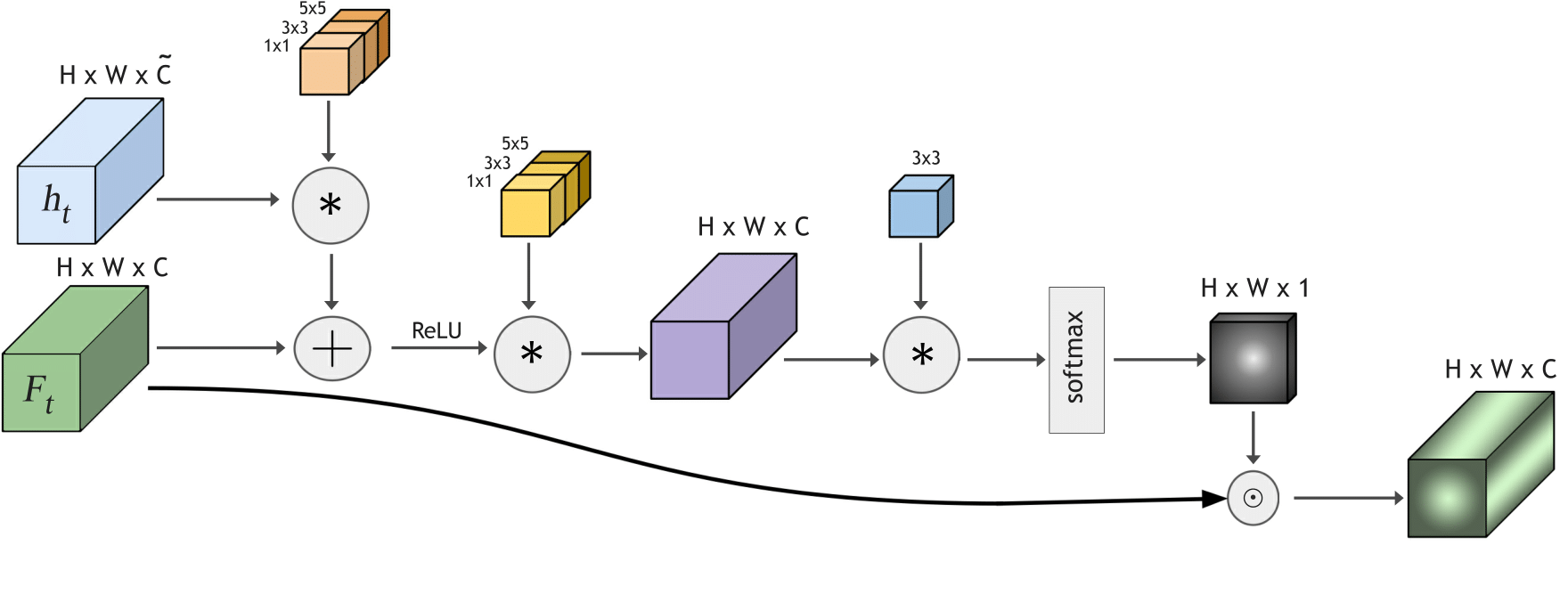}
\caption{Soft Attention Mechanism.}
\label{fig:attn}
\end{figure}

Figure~\ref{fig:attn}, illustrates the soft attention mechanism. Unlike the soft attention mechanism proposed in Xu \etal \cite{xu2015} our's replace fully-connected layers with convolutional layers. Specifically, we used multi-convolutional layers that uses different kernel sizes similar to an inception module. At each time step $t$, the module receives $\mathrm{h}_t$ from ConvLSTM and the selected feature layer $\mathrm{F}_t$. The ConvLSTMs hidden state $\mathrm{h}_t$ is first converted to the appropriate channel size of the feature map. We add the embedding $\mathrm{h}_t$ and  feature layer $\mathrm{F}_t$. Then we apply a non-linearity (Leaky ReLU). After which we compute the attention weights and apply softmax to get the attention map. Then an element-wise multiplication is performed between features and attention map to get the final output of the soft attention module. The Multi-ConvLSTM is applied to attention output. At each time step the LSTM output is used for prediction. In Section~\label{multiapproach} suggest convolutional attention and LSTMs yield better results. We did try using fully-connected LSTMs; however, the system consistently failed to pick different locations in the image during successive LSTM steps.

\section{Datasets} \label{datasets}

\begin{figure*}[h]
   \begin{center}
   \includegraphics[width=\textwidth]{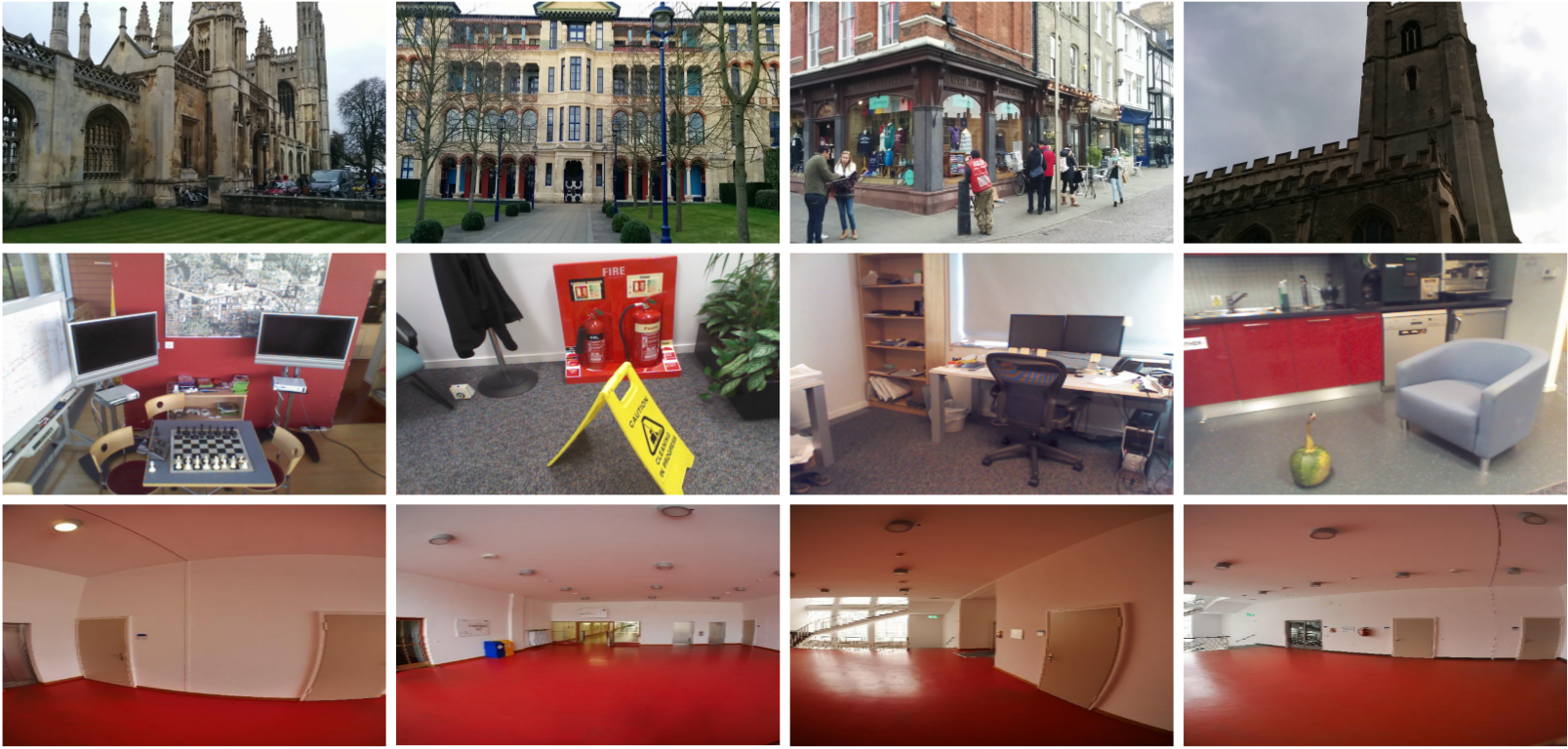}
   \end{center}
   \caption{(a) Top row: Cambridge Landmarks Dataset.  King's College, Old Hospital, Shop Facade and St. Mary's Church. (b) Middle row: 7-Scenes (subset). Chess, Fire, Office and Pumpkin. (c) Bottom row: TUM-LSI.}
   \label{fig:datasets}
\end{figure*}

\begin{figure*}[t]
   \begin{center}
   \includegraphics[width=\textwidth]{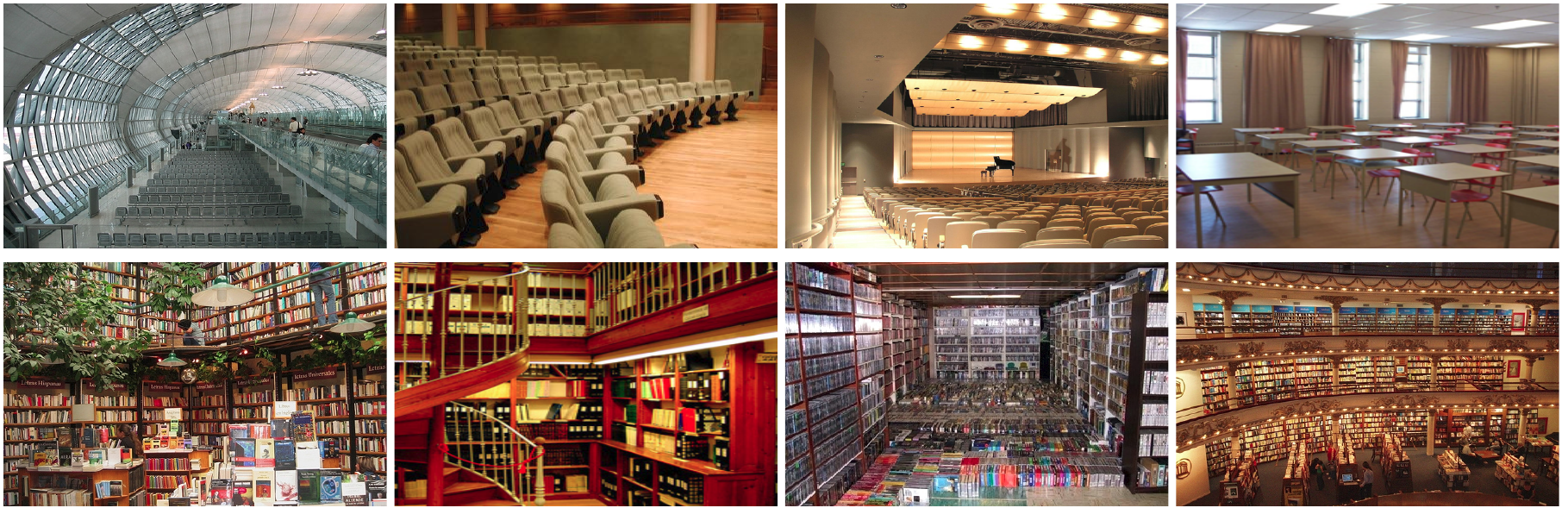}
   \end{center}
   \caption{MIT-67 Indoor Scene Dataset. (a) Top row: Airport, Auditorium, Concert Hall and Classroom. A network can have a hard time classifying them by just focusing on specific properties, since all of them contain large hallways with chairs. (b) Bottom row: Bookstore, Library, Video Store and Library. This set of images have almost the same structure and objects which makes these scenes very ambiguous.}
   \label{fig:datasets}
\end{figure*}

\noindent {\bf Cambridge Landmarks \cite{kendall2015}} A large scale outdoor dataset, containing five outdoor datasets. For our experiments, we only use the four datasets that were used by \cite{kendall2015} and \cite{walch2017}.  The dataset consists of RGB images.  Six degrees-of-freedom camera poses are provided for each image. The dataset was collected using a smart phone, and structure from motion was employed to label each image with its corresponding camera pose.

\noindent {\bf 7-Scenes \cite{shotton2013scene}} A small scale indoor dataset, which consists of seven different scenes.  These scenes were obtained using Kinect RGB-D camera, and KinectFusion\cite{izadi2011kinectfusion} was used to obtain the ground truth. We use the train/test split used by \cite{kendall2015} and \cite{walch2017}. Scene contain ambiguous regions, which makes camera localization difficult.

\noindent {\bf TU Munich Large-Scale Indoor (TUM-LSI) \cite{walch2017}} An indoor dataset, which covers an area of two orders of magnitude larger than that covered by the 7Scenes dataset.  It consists of $875$ training images and $220$ testing images.  We use the train/test split used by \cite{walch2017}.  This is a challenging dataset to localize due to repeated structural elements with nearly identical appearance.

\noindent {\bf MIT-67 indoor scenes \cite{quattoni2009recognizing}} Images taken primarily in four different indoor environments---store, home, public spaces, leisure and working places.  The dataset contains 67 categories in total. We used the official train/test split provided by \cite{quattoni2009recognizing}. Each category has $80$ training images and $20$ testing images.

\section{Extended Implementation details}

\noindent Similar to \cite{kendall2015} and \cite{walch2017}, separate mean images were computed for each channel and the images were mean subtracted per channel.  For Cambridge Landmarks dataset $\beta$ value was set between $250$ to $2000$. For 7-Scenes dataset $\beta$ value was set between $120$ to $750$, and for TUM-LSI dataset $\beta$ value was set to $1000$. For indoor scene classification we mean subtract the Places dataset image mean. For both camera pose estimation and indoor scene classification, we used the same pre-trained CNN layers as used by previous methods. We used the original GoogLeNet weights trained on Places\footnote{\scriptsize \url{http://places.csail.mit.edu/downloadCNN.html}} \cite{zhou2014learning}. We converted these provided trained network weights to be able to use these in \emph{TensorFlow}. The batch size during training was set to $40$. The intial memory states of the LSTM (Memory state $c_0$ and Hidden state $h_0$) is typically set to zero. Similar to \cite{xu2015}, we learn the the initial states. The ConvLSTM hidden size is set to $96$. 

\subsection{Multi-Convolutional Approach} \label{multiapproach}

\begin{table*}[t]
\begin{center}
\centering
\resizebox{\textwidth}{!}{%
\def\arraystretch{1.5}
\begin{tabular}{l>{\hspace{1pc}}c>{\hspace{1pc}}c|>{\hspace{1pc}}c|>{\hspace{1pc}}c}
\specialrule{.2em}{.1em}{.1em}
\multicolumn{1}{l}{\textbf{Dataset}} & \textbf{\begin{tabular}[c]{@{}c@{}}PoseNet \cite{kendall2015} \end{tabular}} & \textbf{\begin{tabular}[c]{@{}c@{}}LSTM-PoseNet \cite{walch2017} \end{tabular}} & \multicolumn{2}{c}{\textbf{Ours}} \\ \specialrule{.1em}{.05em}{.05em}
&  &  & \textbf{\begin{tabular}[c]{@{}c@{}} Convolutional \\ Spatial Attention \end{tabular}} &  \textbf{\begin{tabular}[c]{@{}c@{}} Improvement \\ (meter, degree) \% \end{tabular}} \\ \specialrule{.1em}{.05em}{.05em}
King's College   & 1.66 m, 4.86$^{\circ}$ & \textbf{0.99 m}, 3.65$^{\circ}$ & 1.39 m, \textbf{2.63}$^{\circ}$ & -27.2,  +27.6 \\
Old Hospital     & 2.62 m, 4.90$^{\circ}$ & \textbf{1.51 m}, 4.29$^{\circ}$ & 3.72 m, \textbf{4.24}$^{\circ}$ & -120.5, +6.9 \\ \specialrule{.1em}{.1em}{.1em}
Office   & 0.48 m, 7.24$^{\circ}$ & \textbf{0.30 m}, 8.08$^{\circ}$ & 0.64 m, \textbf{7.89}$^{\circ}$ & -103.3,+3.2 \\
Stairs   & 0.48 m, 13.1$^{\circ}$ & \textbf{0.40 m}, 13.7$^{\circ}$ & 0.48 m, \textbf{12.8}$^{\circ}$ & -15.0, +6.5 \\ \specialrule{.1em}{.1em}{.1em}
TUM-LSI  & 1.87 m, 6.14$^{\circ}$ & \textbf{1.31 m}, 2.79$^{\circ}$ & 3.93 m, \textbf{2.15}$^{\circ}$ & +16, +22.9 \\ \specialrule{.2em}{.1em}{.1em}
\end{tabular}%
}
\end{center}
\caption{Median localization error achieved by the convolutional attention model on a subset of camera pose estimation datasets: Cambridge Landmarks, 7-Scenes, and TUM-LSI dataset. Bold values indicate the lowest error achieved for each row.}
\label{tbl:pose_conv_results}
\end{table*}
\begin{table*}[h]
\begin{center}
\centering
\resizebox{\textwidth}{!}{%
\def\arraystretch{1.5}
\begin{tabular}{l>{\hspace{1pc}}c>{\hspace{1pc}}c|>{\hspace{1pc}}c|>{\hspace{1pc}}c}
\specialrule{.2em}{.1em}{.1em}
\multicolumn{1}{l}{\textbf{Dataset}} & \textbf{\begin{tabular}[c]{@{}c@{}}PoseNet \cite{kendall2015} \end{tabular}} & \textbf{\begin{tabular}[c]{@{}c@{}}LSTM-PoseNet \cite{walch2017} \end{tabular}} & \multicolumn{2}{c}{\textbf{Ours}} \\ \specialrule{.1em}{.05em}{.05em}
&  &  & \textbf{\begin{tabular}[c]{@{}c@{}} Multi-Conv. \\ Spatial Attention \end{tabular}} &  \textbf{\begin{tabular}[c]{@{}c@{}} Improvement \\ (meter, degree) \% \end{tabular}} \\ \specialrule{.1em}{.05em}{.05em}
King's College   & 1.66 m, 4.86$^{\circ}$ & 0.99 m, \textbf{3.65}$^{\circ}$ & \textbf{0.95} m, 4.11$^{\circ}$ & +4.04, -12.6 \\
Old Hospital     & 2.31 m, 5.38$^{\circ}$ & \textbf{1.51} m, \textbf{4.29}$^{\circ}$ & 1.76 m, 4.44$^{\circ}$ & -16.5, -3.49 \\ \specialrule{.1em}{.1em}{.1em}
Office   & 0.48 m, 7.24$^{\circ}$ & 0.30 m, 8.08$^{\circ}$ & \textbf{0.28} m, \textbf{7.52}$^{\circ}$ & +6.67, +6.93 \\
Stairs   & 0.48 m, 13.1$^{\circ}$ & 0.40 m, 13.7$^{\circ}$ & \textbf{0.32} m, \textbf{12.7}$^{\circ}$ & +20.0, +9.40 \\ \specialrule{.1em}{.1em}{.1em}
TUM-LSI  & 1.87 m, 6.14$^{\circ}$ & 1.31 m, \textbf{2.79}$^{\circ}$ & \textbf{1.12} m, 3.66$^{\circ}$ & +14.5, -2.88 \\ \specialrule{.2em}{.1em}{.1em}                                                    
\end{tabular}%
}
\end{center}
\caption{Median localization error achieved by the multi-convolutional attention model on a subset of camera pose estimation datasets: Cambridge Landmarks, 7-Scenes, and TUM-LSI dataset. Bold values indicate the lowest error achieved for each row.}
\label{tbl:pose_multiconv_results}
\end{table*}

\noindent In this section, we describe our motivation for using the multi-convolutional approach. To showcase how we arrived at the proposed approach, we provide evaluation on all three datasets for the pose estimation. We initially started with the same implementation as Xu \etal \cite{xu2015} for soft attention, by using fully connected layers. The model ended up overfitting the data and showed poor performance on the test set. Also, the network converged to select only a single spatial feature instead of probing through the other spatial features at different LSTM time-steps. Our first solution was converting fully connected layers into fully convolutional layers. The results for this approach on pose estimation is shown in Table \ref{tbl:pose_conv_results}. The results shown is quite far from \cite{walch2017} especially on the position, but interestingly error was close to \cite{kendall2015}.

We found that our model was underfitting the training data. Naively increasing the depth size or kernel size was not showing any significant improvements. Therefore by taking inspiration from the inception module proposed in GoogLeNet \cite{szegedy2015}, we converted each convolutional layer into multi-convolutional layers. We used three convolutional kernels with kernel sizes of $1\mathrm{x}1$, $3\mathrm{x}3$ \& $5\mathrm{x}5$ and stacked their final output together. Similarly, in the case of ConvLSTM, we used four convolutional kernels with kernel sizes of $1\mathrm{x}1$, $3\mathrm{x}3$, $5\mathrm{x}5$ \& $7\mathrm{x}7$. Then stacked their final output together for prediction. This approach helped improve results significantly as shown in Table \ref{tbl:pose_multiconv_results}. After which we applied our contribution of layer selection mechanism to form layer-spatial attention. The final results for pose estimation is shown in Table {\color{red} 1} in the main paper.

\newpage
\section{Extended results}

\subsection{Results for Manual Layer Search}
In this section, we show an extensive list of classes in MIT-67 indoor scene classification dataset. This table is an extension to the Table {\color{red} 3} from the main main paper. This is provided to showcase how different layers of CNN capture distinctive information that can help further improve the result.
\begin{table}[!h]
\begin{center}
\centering
\resizebox{0.5\textwidth}{!}{%
\def\arraystretch{1.5}
\begin{tabular}{l|>{\hspace{1pc}}c>{\hspace{1pc}}c>{\hspace{1pc}}c}
\specialrule{.2em}{.1em}{.1em} 
\textbf{Scene} & \begin{tabular}[c]{@{}c@{}} \textbf{Layer} \\   \textbf{3B} \end{tabular} & \begin{tabular}[c]{@{}c@{}}  \textbf{Layer} \\   \textbf{4E} \end{tabular} & \begin{tabular}[c]{@{}c@{}}  \textbf{Layer} \\   \textbf{5B} \end{tabular} \\ \specialrule{.1em}{.05em}{.05em}
Office         & 33.3 & \textbf{52.3} & 42.8 \\
Library        & \textbf{65.0} & 45.0 & 60.0 \\
Wine Cellar    & 71.4 &  \textbf{76.1} & 61.9 \\
Fastfood Restaurant & 58.8 & \textbf{88.2} & 70.5\\
Operating Room & 47.3 & \textbf{52.6} & 36.8 \\
Train Station  & \textbf{85.0} & 65.0 & 60.0 \\ \specialrule{.2em}{.1em}{.1em}
Airport-inside & 40.0 & 60.0 & \textbf{75.0} \\
Closet         & 77.7 & 88.8 & \textbf{94.4} \\
Game Room      & 45.0 & 75.0 & \textbf{80.0} \\
Garage         & 72.2 & 77.7 & \textbf{94.4} \\
Dining room    & 38.8 & 66.6 & \textbf{77.7} \\ 
Locker room    & 66.6 & 85.7 & \textbf{100.0} \\ \specialrule{.2em}{.1em}{.1em}
\end{tabular}%
}
\end{center}
\caption{Indoor scene classification. Mean Accuracy results (\%) after applying spatial soft attention to feature maps from different GoogLeNet layers. Top rows show the classes that improve as we look at different layers. Bottom rows show the classes that decrease performance when looking at other layers. Bold values indicate the highest accuracy achieved for each row.}
\label{tbl:manual_layer_indoor}
\end{table}

\newpage
\subsection{Results for five Conv-LSTM steps}

\begin{table*}[h]
\begin{center}
\centering
\resizebox{\textwidth}{!}{%
\def\arraystretch{1.5}
\begin{tabular}{l>{\hspace{1pc}}c>{\hspace{1pc}}c>{\hspace{1pc}}c>{\hspace{1pc}}c|>{\hspace{1pc}}c>{\hspace{1pc}}c>{\hspace{1pc}}c>{\hspace{1pc}}c>{\hspace{1pc}}c|>{\hspace{1pc}}c}
\specialrule{.2em}{.1em}{.1em}
\multicolumn{1}{l}{\multirow{3}{*}{Dataset}} & \multicolumn{1}{c}{\multirow{3}{*}{\begin{tabular}[c]{@{}c@{}}Area or\\ Volume\end{tabular}}} & \multicolumn{1}{c}{\multirow{3}{*}{\begin{tabular}[c]{@{}c@{}}PoseNet\\ \cite{kendall2015}\end{tabular}}} & \multicolumn{1}{c}{\multirow{3}{*}{\begin{tabular}[c]{@{}c@{}}Bayesian\\ PoseNet \cite{kendall2017uncertainties}\end{tabular}}} & \multicolumn{1}{c}{\multirow{3}{*}{\begin{tabular}[c]{@{}c@{}}LSTM\\ PoseNet \cite{walch2017}\end{tabular}}} & \multicolumn{4}{|c}{Ours} \\ \cline{6-11}
\multicolumn{1}{c}{} & \multicolumn{1}{c}{} & \multicolumn{1}{c}{} & \multicolumn{1}{c}{} & \multicolumn{1}{c|}{} & \begin{tabular}[c]{@{}c@{}}Conv-LSTM \\ Step-1\end{tabular} & \begin{tabular}[c]{@{}c@{}}Conv-LSTM \\ Step-2\end{tabular} & \begin{tabular}[c]{@{}c@{}}Conv-LSTM\\ Step-3\end{tabular} & \begin{tabular}[c]{@{}c@{}}Conv-LSTM\\ Step-4\end{tabular} & \begin{tabular}[c]{@{}c@{}}Conv-LSTM\\ Step-5\end{tabular} & \begin{tabular}[c]{@{}l@{}}Improvement\\ (meter, degree)\end{tabular} \\ \specialrule{.1em}{.05em}{.05em}

Old Hospital  & 2000 $m^2$ & 2.62 m, 4.90$^{\circ}$ & 2.57 m, 5.14$^{\circ}$ & 1.51 m, 4.29$^{\circ}$ & 1.62 m, 4.11$^{\circ}$ & 1.51 m, 4.02$^{\circ}$ & \textbf{1.36} m, \textbf{3.95}$^{\circ}$ & 1.55 m, 4.46$^{\circ}$ & 1.64 m, 4.20$^{\circ}$ & +9.93, +7.92 \\ 

St. Marys Church & 4800 $m^2$ & 2.45 m, 7.96$^{\circ}$ & 2.11 m, 8.38$^{\circ}$ & 1.52 m, 6.68$^{\circ}$ & 1.62 m, 7.22$^{\circ}$ & 1.59 m, 5.94$^{\circ}$ & \textbf{1.42} m, \textbf{6.07}$^{\circ}$  & 1.49 m, 5.87$^{\circ}$ & 1.58 m, 6.51 $^{\circ}$ & +6.57, +1.64 \\ \specialrule{.2em}{.1em}{.1em}
Office      & 7.5 $m^3$ & 0.48 m, 7.24$^{\circ}$ & 0.48 m, 8.04$^{\circ}$ & 0.30 m, 8.08$^{\circ}$ & 0.29 m, 7.63$^{\circ}$ & 0.29 m, 7.23$^{\circ}$ & \textbf{0.29} m, \textbf{8.02}$^{\circ}$ &  0.29 m, 8.07$^{\circ}$ & 0.30 m, 8.12 $^{\circ}$  & +3.33, +0.74 \\
Stairs      & 7.5 $m^3$ & 0.48 m, 13.1$^{\circ}$ & 0.48 m, 13.1$^{\circ}$ & 0.40 m, 13.7$^{\circ}$ & 0.32 m, 9.98$^{\circ}$ & 0.31 m, 10.5$^{\circ}$ & \textbf{0.29} m, \textbf{12.0}$^{\circ}$ & 0.31 m, 12.0$^{\circ}$ & 0.33 m, 10.9 $^{\circ}$  & +27.5, +12.4 \\ \specialrule{.2em}{.1em}{.1em}
TUM-LSI & 5575 $m^2$ & 1.87 m, 6.14$^{\circ}$ & - &1.31 m, 2.79$^{\circ}$ & 1.32 m, 3.82$^{\circ}$ & 1.26 m, 3.69$^{\circ}$ & \textbf{0.98} m, \textbf{2.74}$^{\circ}$ & 1.14 m, 3.33$^{\circ}$ & 1.18 m, 3.68 $^{\circ}$ & +25.1, +1.79 \\ \specialrule{.2em}{.1em}{.1em}
\end{tabular}%
}
\end{center}
\caption{Median localization error achieved by our proposed attention model over five-time steps on subset of Cambridge Landmarks, subset of 7-Scenes, and TUM-LSI. Bold values indicate the lowest error achieved for each row. Improvement is reported with respect to LSTM-PoseNet \cite{walch2017}.}
\label{tbl:pose-results}
\end{table*}

\begin{table*}[h]
\vspace{-4mm}
\begin{center}
\centering
\resizebox{\textwidth}{!}{%
\def\arraystretch{1.5}
\begin{tabular}{c>{\hspace{1pc}}c>{\hspace{1pc}}c|>{\hspace{1pc}}c>{\hspace{1pc}}c>{\hspace{1pc}}c>{\hspace{1pc}}c>{\hspace{1pc}}c|>{\hspace{1pc}}c} \\ \specialrule{.15em}{.1em}{.1em}
\multirow{3}{*}{\begin{tabular}[c]{@{}c@{}}CNNaug-SVM \cite{sharif2014cnn} \end{tabular}} & \multirow{3}{*}{\begin{tabular}[c]{@{}c@{}}S$^2$ICA \cite{hayat2016spatial}\end{tabular}} &
\multirow{3}{*}{\begin{tabular}[c]{@{}c@{}}GoogLeNet \cite{szegedy2015} \end{tabular}} & \multicolumn{4}{c}{\textbf{Ours}} \\ \cline{4-9}
   &  &  & \begin{tabular}[c]{@{}c@{}}Conv-LSTM \\ Step-1\end{tabular} & \begin{tabular}[c]{@{}c@{}}Conv-LSTM \\ Step-2\end{tabular} & \begin{tabular}[c]{@{}c@{}}Conv-LSTM\\ Step-3\end{tabular} & \begin{tabular}[c]{@{}c@{}}Conv-LSTM\\ Step-4\end{tabular} & \begin{tabular}[c]{@{}c@{}}Conv-LSTM\\ Step-5\end{tabular} &\begin{tabular}[c]{@{}c@{}} Improvement (\%) \end{tabular} \\ 
   \specialrule{.1em}{.05em}{.05em}
  69.0 \% & 71.2 \% & 73.7 \% & 74.5 \% & \textbf{77.1} \% & 76.0 \% & 75.4 & 74.8 & +3.4 \\ \specialrule{.15em}{.1em}{.1em}
\end{tabular}%
}
\end{center}
\caption{Mean accuracy results for indoor scene classification on MIT-67. The proposed method achieves the highest accuracy (shown in boldface). Improvement is reported with respect to the GoogLeNet \cite{szegedy2015} baseline.}
\label{tbl:scene-results}
\end{table*}

\noindent \textbf{Camera localization.} We did an experimental study for a subset of scenes from camera localization dataset shown in Table ~\ref{tbl:pose-results}. We concluded that for the camera position estimation Conv-LSTM step three on average provides the best result.
\newline
\newline
\noindent \textbf{Indoor Scene Classification.} We did an experimental study on MIT-67 indoor scene, shown in Table ~\ref{tbl:scene-results}. We concluded that for the Indoor Scene Conv-LSTM step two on average provides the best result.

\end{document}

%% file: intro.tex
\begin{figure*}[t]
   \begin{center}
   \includegraphics[width=0.95\textwidth]{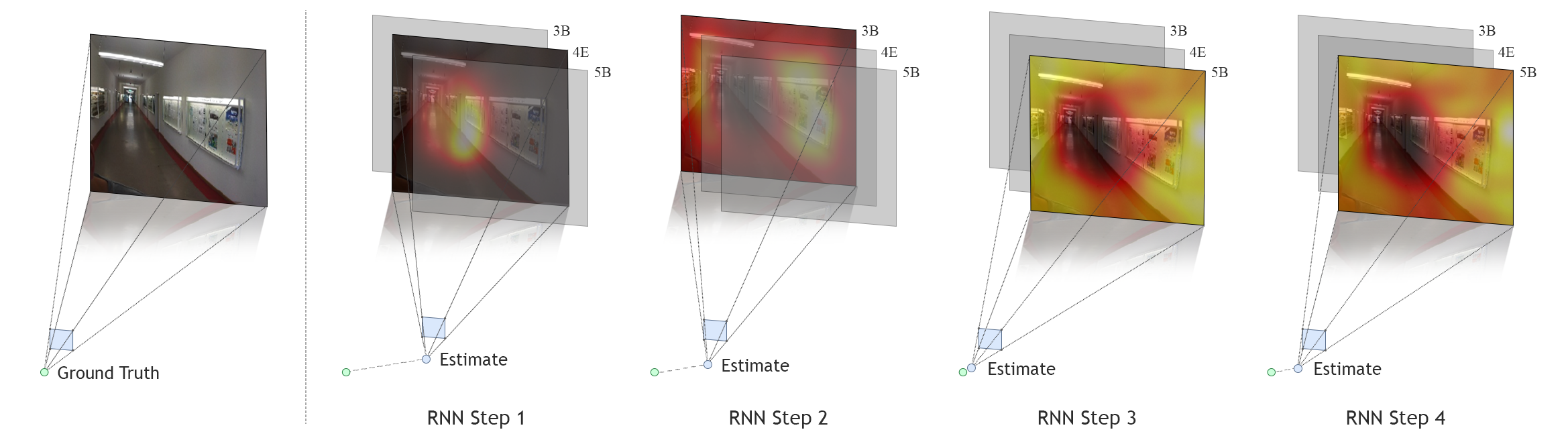}
   \end{center}
\caption{ Overview of our approach to 6-DoF camera localization.  Given a set of CNN feature layers  (GoogLeNet \cite{szegedy2015} Conv-\{3B, 4C, 4E, 5B\} layers  shown) our approach to  attention uses an RNN to sequentially select a set of feature layers (highlighted by the non-grey images) and corresponding locations in the layers (highlighted by the heat maps).  Finally, the processed attended features are used for regressing the camera position and orientation.}
\label{fig:overview}
\end{figure*}

%% file: camera-pose-table.tex
\begin{table*}[t]
\begin{center}
\centering
\resizebox{\textwidth}{!}{%
\def\arraystretch{1.5}
\begin{tabular}{l>{\hspace{1pc}}c>{\hspace{1pc}}c>{\hspace{1pc}}c>{\hspace{1pc}}c|>{\hspace{1pc}}c>{\hspace{1pc}}c>{\hspace{1pc}}c>{\hspace{1pc}}c}
\specialrule{.2em}{.1em}{.1em}
\multicolumn{1}{c}{\multirow{3}{*}{Dataset}} & \multicolumn{1}{c}{\multirow{3}{*}{\begin{tabular}[c]{@{}c@{}}Area or\\ Volume\end{tabular}}} & \multicolumn{1}{c}{\multirow{3}{*}{\begin{tabular}[c]{@{}c@{}}PoseNet\\ \cite{kendall2015}\end{tabular}}} & \multicolumn{1}{c}{\multirow{3}{*}{\begin{tabular}[c]{@{}c@{}}Bayesian\\ PoseNet \cite{kendall2017uncertainties}\end{tabular}}} & \multicolumn{1}{c|}{\multirow{3}{*}{\begin{tabular}[c]{@{}c@{}} LSTM \\ PoseNet \cite{walch2017}\end{tabular}}} & \multicolumn{4}{c}{Ours} \\ \cline{6-9}
\multicolumn{1}{c}{} & \multicolumn{1}{c}{} & \multicolumn{1}{c}{} & \multicolumn{1}{c}{} & \multicolumn{1}{c|}{} & \begin{tabular}[c]{@{}c@{}}Conv-LSTM \\ Step-1 \end{tabular} & \begin{tabular}[c]{@{}c@{}}Conv-LSTM \\ Step-2\end{tabular} & \begin{tabular}[c]{@{}c@{}} Conv-LSTM \\ Step-3 \end{tabular} & \begin{tabular}[c]{@{}c@{}}Improvement \\ (meter, degree)\end{tabular} \\ \specialrule{.1em}{.05em}{.05em}

Great Court   & 8000 $m^2$ &  - & - & - & - & - & - & - \\
Kings College & 5600 $m^2$ & 1.66 m, 4.86$^{\circ}$ & 1.74 m, 4.06$^{\circ}$ & 0.99 m, \textbf{3.65}$^{\circ}$ & 1.02 m, 4.22$^{\circ}$ & 1.00 m, 4.51$^{\circ}$ & \textbf{0.90} m, 3.70$^{\circ}$  & +9.09, -1.36 \\
Old Hospital  & 2000 $m^2$ & 2.62 m, 4.90$^{\circ}$ & 2.57 m, 5.14$^{\circ}$ & 1.51 m, 4.29$^{\circ}$ & 1.62 m, 4.11$^{\circ}$ & 1.51 m, 4.02$^{\circ}$ & \textbf{1.36} m, \textbf{3.95}$^{\circ}$ & +9.93, +7.92\\
Shop Facade   & 875 $m^2$  & 1.41 m, 7.18$^{\circ}$ & 1.25 m, 7.54$^{\circ}$ & 1.18 m, 7.44$^{\circ}$ & 1.15 m, 5.45$^{\circ}$ & 0.95 m, 6.44$^{\circ}$ & \textbf{0.91} m, \textbf{5.29}$^{\circ}$ & +22.8, +28.8 \\
St. Marys Church & 4800 $m^2$ & 2.45 m, 7.96$^{\circ}$ & 2.11 m, 8.38$^{\circ}$ & 1.52 m, 6.68$^{\circ}$ & 1.62 m, 7.22$^{\circ}$ & 1.59 m, 5.94$^{\circ}$ & \textbf{1.42} m, \textbf{6.07}$^{\circ}$ & +6.57, +1.64 \\
Street        & 50000 $m^2$ &  - & - & - & 18.7m, 34.1$^{\circ}$ & 15.0 m, 30.3$^{\circ}$ & \textbf{13.9} m, \textbf{30.0}$^{\circ}$ & - \\ \specialrule{.1em}{.05em}{.05em}
Average \cite{walch2017} & 3319 $m^2$ & 2.08 m, 6.83$^{\circ}$ & 1.92 m, 6.28$^{\circ}$ & 1.30 m, 5.52$^{\circ}$ & 1.35 m, 5.25$^{\circ}$ & 1.26 m, 5.22$^{\circ}$ & \textbf{1.14} m, \textbf{4.75}$^{\circ}$ & +12.3, +13.9 \\ \specialrule{.2em}{.1em}{.1em}
Chess       & 6.0 $m^3$  & 0.32 m, 6.08$^{\circ}$ & 0.37 m, 7.24$^{\circ}$ & 0.24 m, 5.77$^{\circ}$ & 0.17 m, 5.58$^{\circ}$ & 0.16 m, 5.27$^{\circ}$ & \textbf{0.15} m, \textbf{4.79}$^{\circ}$ & +37.5, +16.9\\
Fire        & 2.5 $m^3$ & 0.47 m, 14.0$^{\circ}$ & 0.43 m, 13.7$^{\circ}$ & 0.34 m, 11.9$^{\circ}$ & 0.32 m, 12.6$^{\circ}$ & 0.31 m, 11.7$^{\circ}$ & \textbf{0.23} m, \textbf{10.0}$^{\circ}$ & +32.3, +15.9 \\
Heads       & 1.0 $m^3$   & 0.30 m, 12.2$^{\circ}$ & 0.31 m, 12.0$^{\circ}$ & 0.21 m, 13.7$^{\circ}$ & 0.18 m, 13.8$^{\circ}$ & 0.18 m, 14.1$^{\circ}$ & \textbf{0.18} m, \textbf{13.7}$^{\circ}$ & +14.2, +0.00 \\
Office      & 7.5 $m^3$ & 0.48 m, 7.24$^{\circ}$ & 0.48 m, 8.04$^{\circ}$ & 0.30 m, 8.08$^{\circ}$ & 0.29 m, 7.63$^{\circ}$ & 0.29 m, 7.23$^{\circ}$ & \textbf{0.29} m, \textbf{8.02}$^{\circ}$ & +3.33, +0.74 \\
Pumpkin     & 5.0 $m^3$   & 0.49 m, 8.12$^{\circ}$ & 0.61 m, 7.08$^{\circ}$ & 0.33 m, 7.00$^{\circ}$ & 0.25 m, 5.46$^{\circ}$ & 0.25 m, 5.76$^{\circ}$ & \textbf{0.26} m, \textbf{6.16}$^{\circ}$ & +21.2, +12.0 \\ 
Red Kitchen & 18 $m^3$  & 0.58 m, 8.31$^{\circ}$ & 0.58 m, 7.51$^{\circ}$ & \textbf{0.37} m, 8.83$^{\circ}$ & 0.43 m, 8.03$^{\circ}$ & 0.37 m, 7.49$^{\circ}$ & 0.39 m, \textbf{8.20}$^{\circ}$ & -2.00, +5.77  \\
Stairs      & 7.5 $m^3$ & 0.48 m, 13.1$^{\circ}$ & 0.48 m, 13.1$^{\circ}$ & 0.40 m, 13.7$^{\circ}$ & 0.32 m, 9.98$^{\circ}$ & 0.31 m, 10.5$^{\circ}$ & \textbf{0.29} m, \textbf{12.0}$^{\circ}$ & +27.5, +12.4 \\ \specialrule{.1em}{.05em}{.05em}
Average All & 6.9 $m^3$ &  0.44 m, 9.01$^{\circ}$ & 0.46 m, 9.81$^{\circ}$ & 0.31 m, 9.85$^{\circ}$ & 0.28 m, 9.01$^{\circ}$ & 0.26 m, 8.86$^{\circ}$ & \textbf{0.25} m, \textbf{8.98}$^{\circ}$ & +19.1, +9.10 \\ \specialrule{.2em}{.1em}{.1em}
TUM-LSI & 5575 $m^2$ & 1.87 m, 6.14$^{\circ}$ & - &1.31 m, 2.79$^{\circ}$ & 1.32 m, 3.82$^{\circ}$ & 1.26 m, 3.69$^{\circ}$ & \textbf{0.98} m, \textbf{2.74}$^{\circ}$  & +25.1, +1.79 \\ \specialrule{.2em}{.1em}{.1em}
\end{tabular}%
}
\end{center}
\caption{Camera localization results. Median localization error achieved by the proposed attention model over three steps on Cambridge Landmarks, 7-Scenes, and TUM-LSI. Bold values indicate the lowest error achieved for each row. Improvement is reported with respect to LSTM-PoseNet \cite{walch2017}. A dash (-) indicates that no result is reported. }
\label{tbl:pose-results-cam}
\end{table*}

%% file: scene-classification-table.tex
\begin{table*}[t]
\vspace{-4mm}
\begin{center}
\centering
\resizebox{\textwidth}{!}{%
\def\arraystretch{1.5}
\begin{tabular}{c>{\hspace{1pc}}c>{\hspace{1pc}}c|>{\hspace{1pc}}c>{\hspace{1pc}}c>{\hspace{1pc}}c|>{\hspace{1pc}}c>{\hspace{1pc}}c>{\hspace{1pc}}c} \\ \specialrule{.15em}{.1em}{.1em}
\multirow{3}{*}{\begin{tabular}[c]{@{}c@{}}CNNaug-SVM \cite{sharif2014cnn} \end{tabular}} & \multirow{3}{*}{\begin{tabular}[c]{@{}c@{}}S$^2$ICA \cite{hayat2016spatial}\end{tabular}} &
\multirow{3}{*}{\begin{tabular}[c]{@{}c@{}}GoogLeNet \cite{szegedy2015} \end{tabular}} & \multicolumn{4}{c}{\textbf{Ours}} \\ \cline{4-7}
   &  &  & \begin{tabular}[c]{@{}c@{}}Conv-LSTM \\ Step-1\end{tabular} & \begin{tabular}[c]{@{}c@{}}Conv-LSTM \\ Step-2\end{tabular} & \begin{tabular}[c]{@{}c@{}}Conv-LSTM\\ Step-3\end{tabular} & \begin{tabular}[c]{@{}c@{}} Improvement (\%) \end{tabular} \\ 
   \specialrule{.1em}{.05em}{.05em}
  69.0 \% & 71.2 \% & 73.7 \% & 74.5 \% & \textbf{77.1} \% & 76.0 \% & +3.4 \\ \specialrule{.15em}{.1em}{.1em}
\end{tabular}%
}
\end{center}
\caption{Mean accuracy results for indoor scene classification on MIT-67. The proposed method achieves the highest accuracy (shown in boldface). Improvement is reported with respect to the GoogLeNet \cite{szegedy2015} baseline.}
\label{tbl:scene-results}
\end{table*}

%% file: ablation-study-table.tex
\begin{table*}[t]
\begin{center}
\centering
\resizebox{\textwidth}{!}{%
\def\arraystretch{1.5}
\begin{tabular}{l|>{\hspace{1pc}}c|>{\hspace{1pc}}c|>{\hspace{1pc}}c|>{\hspace{1pc}}c|>{\hspace{1pc}}c}
\hline
\multirow{2}{*}{Dataset}
 & \multicolumn{3}{c|}{Spatial Attention Only} & 
 \multirow{2}{*}{Layer  Selection Only}
  & \multirow{2}{*}{\begin{tabular}[c]{@{}c@{}}Spatial and Layer Attention\end{tabular}} \\ \cline{2-4}
 & Conv-3B & Conv-4E & Conv-5B &  &  \\ \hline
\multicolumn{6}{c}{Camera-Pose Estimation} \\ \hline
Old Hospital & 1.49 m, 4.29$^{\circ}$ & 1.42 m, 4.37$^{\circ}$ &  1.76 m, 4.44$^{\circ}$ & 2.36 m, 6.28$^{\circ}$ & \textbf{1.36} m, \textbf{3.95}$^{\circ}$\\ \hline
Office       & 0.27 m, 7.37$^{\circ}$ & \textbf{0.26} m, \textbf{7.35}$^{\circ}$ &  0.28 m, 7.52$^{\circ}$  & 0.33 m, 7.97$^{\circ}$ &  0.29 m, 8.02$^{\circ}$\\ \hline
TUM-LSI      & 1.21 m, 3.26$^{\circ}$ & 1.13 m, 3.66$^{\circ}$ &  1.12 m, 3.66$^{\circ}$ & 5.27 m, 10.8$^{\circ}$ & \textbf{0.98} m, \textbf{2.74}$^{\circ}$\\ \hline
\multicolumn{6}{c}{Indoor-Scene Classification} \\ \hline
MIT-67  & 61.6 \% & 74.5 \% & 74.2 \% & 76.4 \%  & \textbf{77.1} \% \\ \hline
\end{tabular}%
}
\end{center}
\caption{Ablation study on layer-spatial attention. In all cases, GoogLeNet \cite{szegedy2015} Conv-\{3B, 4E, 5B\} layers are used. Bold values indicate the best result achieved for each row.}
\label{tab:ablation}
\end{table*}